\documentclass[10pt,twocolumn,letterpaper]{article}

\usepackage{wacv}
\usepackage{times}
\usepackage{epsfig}
\usepackage{graphicx}
\usepackage{amsmath}
\usepackage{amssymb}

\usepackage{booktabs}
\usepackage{multirow}

\def\wacvPaperID{0523} 

\wacvfinalcopy 

\ifwacvfinal
\usepackage[breaklinks=true,bookmarks=false]{hyperref}
\else
\usepackage[pagebackref=true,breaklinks=true,colorlinks,bookmarks=false]{hyperref}
\fi

\begin{document}

\title{Coarse- and Fine-grained Attention Network with Background-aware Loss for Crowd Density Map Estimation}

\author{Liangzi Rong \hspace{0.5cm} Chunping Li\\
School of Software, Tsinghua University\\
{\tt\small lzrong13@gmail.com \hspace{0.5cm} cli@tsinghua.edu.cn}
}

\maketitle
\pagestyle{empty}
\thispagestyle{empty}

\begin{abstract}
In this paper, we present a novel method Coarse- and Fine-grained Attention Network (CFANet) for generating high-quality crowd density maps and people count estimation by incorporating attention maps to better focus on the crowd area. We devise a from-coarse-to-fine progressive attention mechanism by integrating Crowd Region Recognizer (CRR) and Density Level Estimator (DLE) branch, which can suppress the influence of irrelevant background and assign attention weights according to the crowd density levels, because generating accurate fine-grained attention maps directly is normally difficult. We also employ a multi-level supervision mechanism to assist the backpropagation of gradient and reduce overfitting. Besides, we propose a Background-aware Structural Loss (BSL) to reduce the false recognition ratio while improving the structural similarity to groundtruth. Extensive experiments on commonly used datasets show that our method can not only outperform previous state-of-the-art methods in terms of count accuracy but also improve the image quality of density maps as well as reduce the false recognition ratio.
\end{abstract}

\section{Introduction}

Recently, crowd density map estimation has received continuous attention as a challenging computer vision task. Given a crowd image, its purpose is to estimate the density map and the total number of people. The value of each pixel in the density map reflects the density of the corresponding area in the image, and estimated people count can be obtained by accumulating the values of all pixels. Generally speaking, there are three major difficulties for accurate crowd counting: (1) Different distances from the shooting device lead to scale variation within one image and between different images. (2) Severe occlusion in high-density crowd scenes. (3) The influence of complex and irrelevant background is not conducive to recognizing crowd areas.
\begin{figure}[t]
    \centering
    \includegraphics[width=\linewidth]{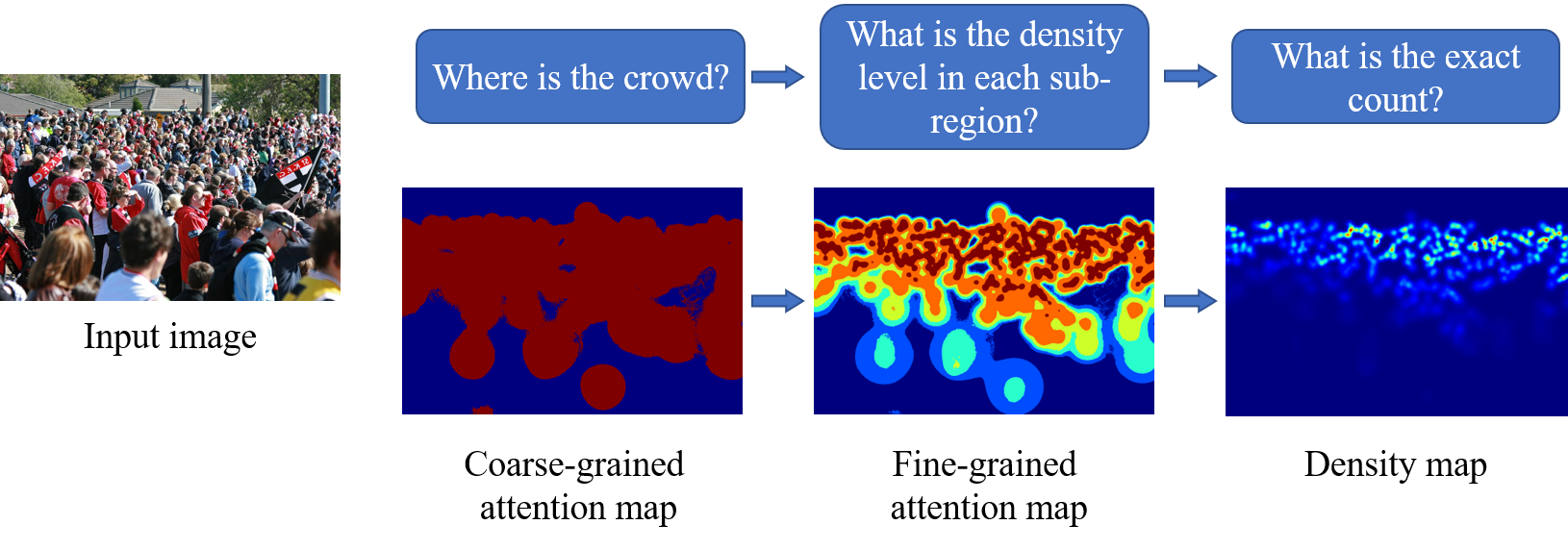}
    \caption{An intuitive explanation of our method.}
    \label{fig:idea}
\end{figure}
Existing methods\cite{cao2018scale, zhang2019relational, liu2019crowd, liu2019adcrowdnet} mainly adopt multi-scale or multi-column architecture-based CNN models to capture richer features. But they generally have two disadvantages: (1) The influence of the background is not fully considered. Existing methods treat all the regions in one image as potential crowd areas, so the convolution kernels may extract features in the background regions that are not related to the crowd, resulting in misrecognition. In this case, even if the estimated number of people across the whole image is close to the ground truth, it may be caused by both the underestimation of the crowd region and the misrecognition of the background region. (2) All crowd regions are treated equally. In fact, when a person observes a crowd image, it is normal to pay different attention to different areas. This is because for areas with a high degree of occlusion, it is more difficult to distinguish the features of each person, thus worth more attention. In low-density areas, it is easier to distinguish each person, thus worth less attention. However, most existing methods pay the same attention to all sub-regions with in one image.

In order to solve the above two problems, we hope to design an attention mechanism that can suppress the influence of the background and reduce misrecognition, but also adaptively assign attention weights to regions with different density. Therefore, we introduce the attention maps. We expect that in one attention map, the background area has a weight close to 0, and the low-density area has relatively low weights, and the high-density area has relatively high weights. As depicted in Fig. \ref{fig:idea}, when the human eyes observe a picture, they will first recognize in which areas people exist, and then identify the dense degree and count. Intuitively, it is natural to follow this paradigm when designing a CNN. It is easier to judge whether there are people in an area than to judge the density level of the area, so we can get more reliable results if we start from simple tasks. Therefore, we employ two modules, Crowd Region Recognizer (CRR) and Density Level Estimator (DLE), to judge whether people exist and the density level for an area respectively.

To be more specific, CRR produces a coarse-grained attention map (CAM), whose value indicating the possibility of if people exist. DLE produces a fine-grained attention map (FAM), whose value indicating the density level of each area. In order to leverage the relatively reliable CAM, we combine the FAM with it. Then feature maps for regressing density maps are combined with FAM on multiple scale to pay adaptive attention to different areas.

The images of crowd datasets is relatively few compared to other datasets, e.g. Imagenet. To reduce overfitting and facilitate the backpropagation of gradients, we introduce multi-level supervision by adding several more output layers in internal layers and summing all the loss functions and backpropagate.

In addition, this paper also explores the impact of different loss functions. We propose a novel and effective loss function named Background-aware Structural Loss (BSL) that can take account of structural similarity, counting accuracy and false recognition ratio. We evaluate the performance on multiple datasets and models, and proposed BSL delivers superior performance than other loss functions.

To sum up, our contribution are three-fold: (1) We present Coarse- and Fine-grained Attention Network (CFANet) that can produce high-quality density maps and accurate count estimation. Crowd Region Recognizer (CRR) and Density Level Estimator (DLE) are employed to estimated coarse- and fine-grained attention maps respectively to help pay more attention to areas with high density and less attention to areas without people or with a low density. (2) We introduce the multi-level supervision mechanism to facilitate the backpropagation of gradients and reduce overfitting. (3) We propose a loss function named Background-aware Structural Loss (BSL) that can improve structural similarity and reduce false recognition. By combining CFANet with BSL, we can achieve the best performance on most mainly used datasets.

\section{Related Work}
We classify existing methods into two categories, one is only using the density map as the learning objective, and the other is combining classification, segmentation and other tasks as the learning objectives.

\textbf{Density Map as the Learning Objective.}
To solve the problem that the size of the heads in the crowd scene changes greatly, MCNN\cite{zhang2016single} designed three parallel subnetworks with convolution kernels of different sizes. CSRNet\cite{li2018csrnet} found that the subnetworks introduced redundant parameters, so a single-column model was proposed. SANet\cite{cao2018scale} combined convolution kernels of different sizes as the feature extraction part, which enhances the non-linear expression ability of the network. CAN\cite{liu2019context} used pooling pyramid to extract features of different scales, and adaptively assigns different weights to different scales and regions. TEDNet\cite{jiang2019crowd} proposed a lightweight hierarchical network structure by combining high-level and low-level features effectively. RR\cite{wan2019residual} used representative images in the dataset obtained by clustering to assist regress the residual and final density map. Bayesian loss\cite{ma2019bayesian} was proposed based on the Bayesian probability model. The value of each pixel in the density map contributes to each person's value or the background by probability. L2SM\cite{xu2019learn} further zooms in and re-estimates the dense area to improve the performance in the high-density area after obtaining the initial density map. DSSINet\cite{liu2019crowd} uses conditional random fields to enable the features of each scale to obtain information from other scales for improvement. 

\textbf{Combinal Learning Objectives.}
CP-CNN\cite{sindagi2017generating} uses extra classification networks to combine global and local density level to improve model accuracy. Switch-CNN\cite{sam2017switching} classifies the input images into three categories according to density level, and then processes the images with corresponding sub-network. DecideNet\cite{liu2018decidenet} constructed a model using two sub-networks for detection and regression respectively, and the outputs of the two sub-networks are fused using attention mechanism. RAZNet\cite{liu2019recurrent} uses an additional localization branch to detect the heads' position, and adaptively zooms in the regions that are difficult to recognize. SGANet\cite{wang2019segmentation} employs the Inception-V3 structure as the feature extraction part, and uses the penultimate layer's feature maps to regress an attention map. 

The most similar work to ours is ADCrowdNet\cite{liu2019adcrowdnet} which uses a separate classification network to obtain the attention map, and multiplies it with the image or feature maps. We differ from it from three aspects: (1) We find that single scale combination has limited influence so we combine the attention maps on multiple scales to enhance the attention effect. (2) Attention maps in ADCrowdNet is similar to our coarse attention maps, but we further incorporate the density level information, which is not included in ADCrowdNet. (3) We do not need a separate classification network to produce attention maps, so our method is much easier for end-to-end training.
 \begin{figure*}[ht]
    \centering
    \includegraphics[width=.95\linewidth]{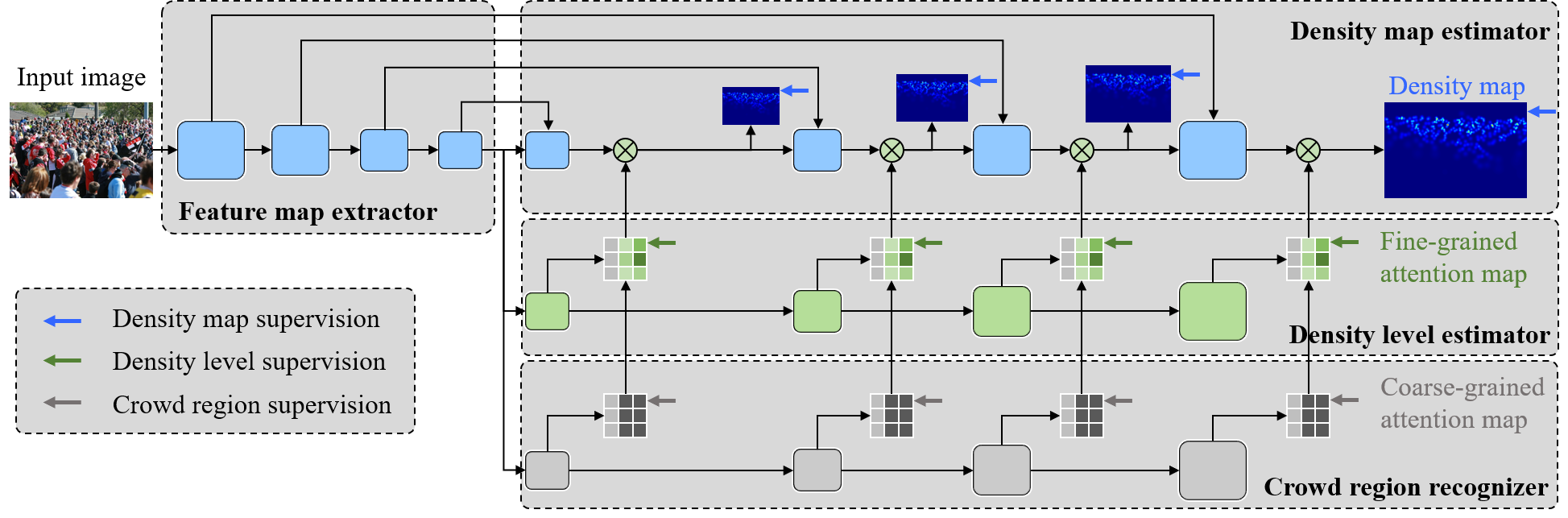}
    \caption{The structure of proposed CFANet.}
    \label{fig:cfanet}
\end{figure*}
\section{Our Approach}
The architecture of the proposed method is illustrated in Figure \ref{fig:cfanet}. It consists of four modules: Feature map extractor is used to extract general feature maps from crowd images and feed them into the following modules for further processing. Crowd region recognizer (CRR) is used to judge whether an area contains crowd, and outputs a coarse-grained attention map (CAM). Density level estimator (DLE) is used to further classify the density level of each area, and can output a fine-grained attention map (FAM). With the aid of FAM, density map estimator can better focus on the crowd region and produce high-quality density maps. In addition, to assist the backpropagation and reduce overfitting, we design a multi-level supervision mechanism. In each stage of the density map estimator, feature maps are upsampled and fed into $Conv$ layers to obtain a density map, and the loss functions of multiple stages are summed and backpropagated.

\subsection{Feature map extractor}
Coarse- and fine-grained attention map and density map, which will be detailed in the following subsections, share a feature: having high values in the high-density area, and low values in the low-density area and near zero in the background area. Therefore, we argue that general features can be extracted through the same base network.

In most existing studies, the resolution of estimated density maps are lower than input, leading to the loss of spatial details. Inspired by the success of UNet\cite{ronneberger2015u}, we design our model in an encoder-decoder paradigm. Following the practice of most previous work, we adopt the VGG-16's feature extraction part in this part. As Fig. \ref{fig:cfanet} shows, we retain the first 10 convolutional layers and 3 pooling layers. Therefore, we can get feature maps with sizes of 1, 1/2, 1/4, and 1/8 from each stage.

\subsection{Crowd region recognizer (CRR)}
Because the crowd images contain different scenes, accurate crowd counting may be hindered by complex backgrounds. Even if the overall estimated number of people is close to the groundtruth, it may be caused by the underestimation of the crowd area and the false recognition of the background area. To address this issue, estimated coarse-grained attention maps (CAM) from CRR try to classify each pixel in feature maps into two categories: crowd and background region. Detailed configuration of the CRR module is: C(256,3)-U-C(128, 3)-U-C(128, 3)-U-C(64, 3)-C(1, 3), where C means convolution layer and U means bilinear upsample layer with rate=2. At each stage in CRR, the feature maps are fed into a 3$\times$3 $Conv$ layer to regress a coarse-grained attention map which is then fed into corresponding stage in DLE. Losses calculated on multi-stage are summed and backpropagated as the gray arrow in Fig. \ref{fig:cfanet}. We will describe how to generate the groundtruth CAM, i.e., learning objective of this module, in Section 4.1.

\begin{figure*}[ht]
    \centering
    \includegraphics[width=.95\linewidth]{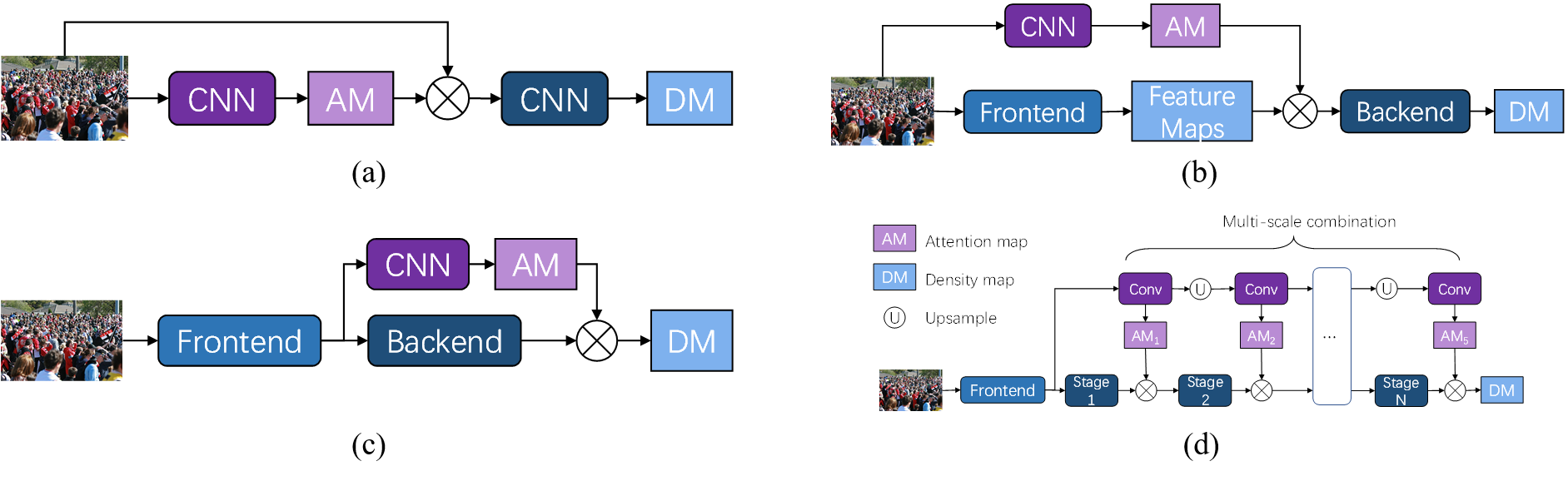}
    \caption{Comparison of attention-based methods. AME means attention map estimator. DME means density maps estimator. Proposed method is (d).}
    \label{fig:attention}
\end{figure*}

\subsection{Density level estimator (DLE)}
The CRR module only implements a coarse-grained attention mechanism, that is, it only distinguishes between people and background. The goal of the DLE module is to further classify the crowd area into different density levels, because for high-density areas, we should give more attention, and for low-density areas, we should give less attention. Similar to CRR, this module should also try to suppress the influence of unmanned background. To this end, we divide all pixels into $k$ categories according to the threshold obtained from statistics and consider it as a $k$-class classification problem. This process of the generating groundtruth FAM, i.e., learning objective of DLE module, is detailed in Section 4.1. After the $k$-class classification, different attention weights are assigned accordingly: pixels classified to class 0 are regarded as the background, and the attention weight are set to 0. For other density levels, we divide the range of (0,1] into k-1 categories to correspond. For example, when k is 6, classes 1-5 correspond to 0.2, 0.4, 0.6, 0.8 and 1 respectively. Regarding the number of categories, we have conducted an experimental comparison, and found that when the number of categories is set to 6, the performance is best. Detailed experimental results can be seen in section 4. Detailed configuration of the DLE module is: C(256,3)-U-C(256, 3)-U-C(128, 3)-U-C(64, 3)-C($k$, 3), where C means convolution layer and U means bilinear upsample layer. At each stage in DLE, feature maps are fed into a 3$\times$3 $Conv$ layer and regress a fine-grained attention map. Similar to CRR, losses calculated on multi-stage are summed and backpropagated as the green arrow in Fig. \ref{fig:cfanet}. Since we already have the CAM from CRR, we can make full use of it to refine the FAM: 
\begin{equation}
    FAM = FAM + CAM
\end{equation}
Then the FAM is fed into the corresponding stage in the next part.

\subsection{Density map estimator}
This module’s configuration is almost symmetrical with the feature map extractor. Since CFANet aims to estimate high-resolution and high-quality density maps, DME is constructed with a set of convolutional and upsample layers: C(512, 3)-U-C(256, 3)-U-C(256, 3)-U-C(64, 3)-C(1, 1), where C is convolution and U bilinear upsample layer. We set dilation rate = 2 in all convolutional layers to enlarge the receptive field. 

There have been studies using the attention map to better focus on the crowd area. As Fig. \ref{fig:attention} shows, (a) and (b) train a separate network to predict the attention map. (a) multiplies the input image with predicted attention map and then use it for regression. (b) multiplies feature maps extracted by the frontend with the predicted attention map. (c) inserts an attention branch to predict attention map before the last convolutional layer. The density map is multiplied with the predicted attention map as the final output. It is worth noting, however, that the weights assigned to the background area by estimated attention map is not equal to 0, so the influence cannot be completely suppressed with one-time combination. Therefore, we adopt a multi-scale attention combination method, as shown in Figure \ref{fig:attention} (d). At each stage in this module, the feature maps (FM) are combined with the fine-grained attention maps (FAM) from the DLE in a residual way:
\begin{equation}
    FM = FM+FAM*FM
\end{equation}
In this way, the attention effects are enhanced and different areas are paid adaptive attention with the aid of fine-grained attention maps.

We also employ a multi-level supervision mechanism. Feature maps in each stage will be upsampled to the size of the input image, and fed into a 3$\times$3 convolutional layer to regress a density map as the blue arrow in Fig. \ref{fig:cfanet} indicates. Therefore, for each input image, CFANet will produce 4 density maps and the 4 losses of are summed and backpropagated. Generally, a deeper network has stronger expression ability, so we regard the output of the last layer as the final output.

\subsection{Loss function}
We first introduce a structural loss function (SL) that considers both structural similarity and counting accuracy. The definition is as follows:
\begin{equation}
    SL=\frac{1}{K}\sum_{j=1}^{K} (1-SSIM(Pool_{j}(DM), Pool_{j}(\hat{DM})))
\end{equation}
\begin{equation}
    SSIM(X,Y)=1-(\frac{\left(2 \mu_{X} \mu_{Y}+C_{1}\right)\left(2 \sigma_{XY}+C_{2}\right)}{\left(\mu_{X}^{2}+\mu_{Y}^{2}+C_{1}\right)\left(\sigma_{X}^{2}+\sigma_{Y}^{2}+C_{2}\right)})
    \label{eq:ssim}
\end{equation}
where $DM$ and $\hat{DM}$ mean groundtruth and estimated density map and $Pool_{j}$ means downsampling to $\frac{1}{2^{j-1}}$ size with average pooling. $\mu$ denotes the local mean and $\sigma$ is the local variance, $\sigma_{XY}$ is the local covariance. $C_1$ and $C_2$ are set to 0.01 and 0.03. $K$ is set to 3. SSIM of high resolution density maps can focus on spatial details and improve structural similarity. SSIM of downsampled density maps can improve global counting accuracy.

To reduce the false recognition ratio, we add a background-aware loss item (BL):
\begin{equation}
    BL=\frac{C_{bg}}{C_{total}}
    \label{eq:bl}
\end{equation}
where $C_{total}$ is the estimated total people count and $C_{bg}$ is the estimated people count in the background area. The background area is divided in the same way as the groundtruth CAM which is detailed in section 4.1.

The loss function (BSL) for density map optimization is the sum of SL and BL.

To optimize the coarse- and fine-grained attention maps, we use cross-entropy as the loss function.

The final loss function is the weighted sum of density map and attention maps loss functions at multiple stages.
\begin{equation}
    L_{total}=SL+BL+\lambda L_{CAM}+\mu L_{FAM}
\end{equation}

\begin{figure}[ht]
    \centering
    \includegraphics[width=\linewidth]{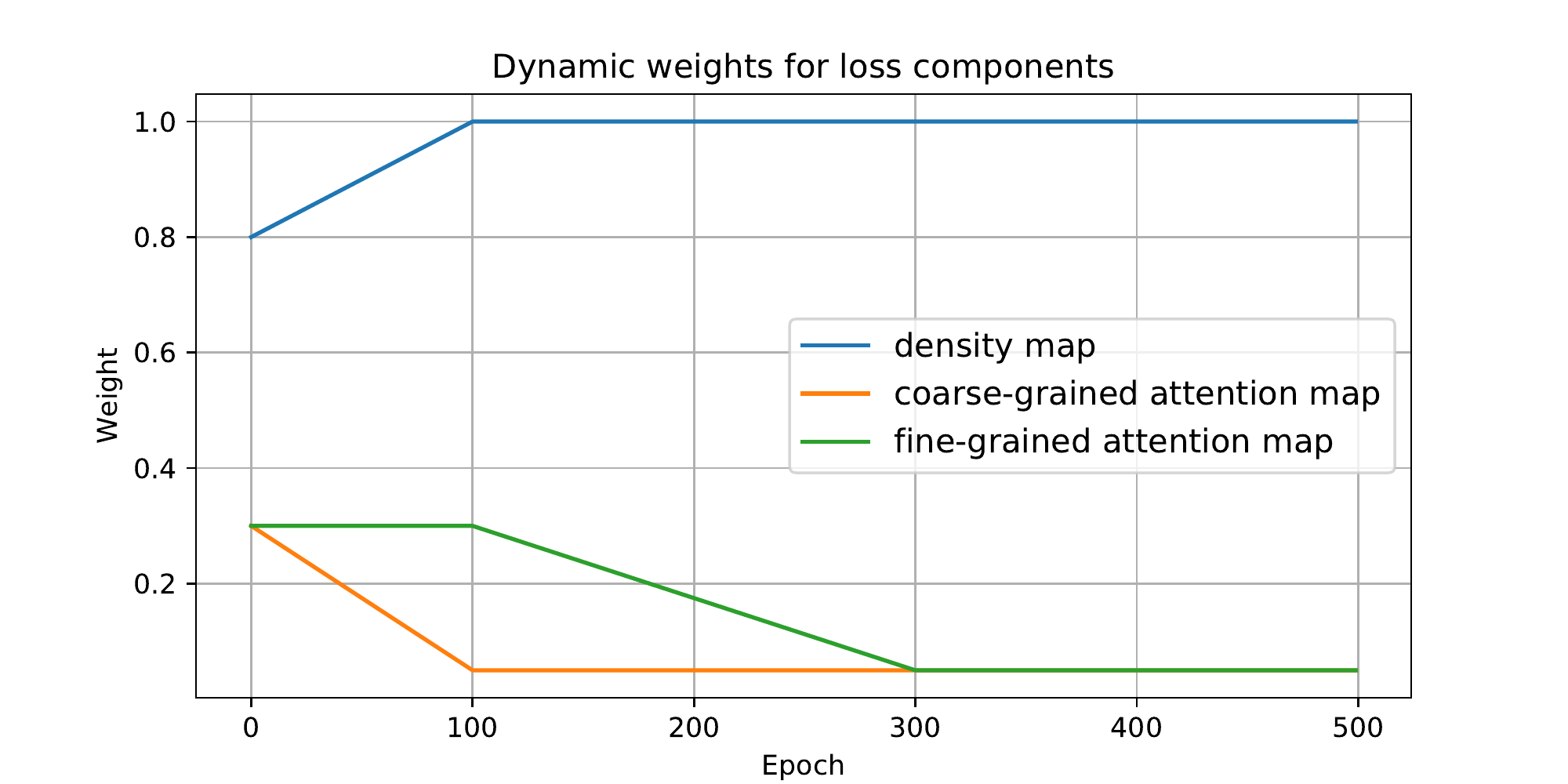}
    \caption{Dynamic weight adjustment strategy.}
    \label{fig:adj}
\end{figure}
Since it is the easiest task to divide the image into background and crowd, and the density multi-classification has medium difficulty, and it is the most difficult to regress the accurate density value of each pixel, we adopt a dynamic weight adjusting strategy in the loss function. In the initial training stage, the network cannot accurately capture the distribution features. Therefore, we focus more on the easy tasks, and set the weights of binary classification and multi-class classification large. As the training process progresses, the representative ability of the network continues to improve, and the features can be better captured. At this time, the weight of density map loss is adjusted to be larger. This process is shown in the Fig. \ref{fig:adj}.

\section{Experiments and Analysis}
\subsection{Datasets and Evaluation Criteria}
We evaluate the performance of proposed CFANet on four major crowd counting datasets: ShanghaiTech \cite{zhang2016single}, UCF\_CC\_50 \cite{idrees2013multi}, UCF-QNRF \cite{idrees2018composition} and Mall \cite{chen2012feature}. The detailed comparisons are described in the following subsections.

\textbf{Groundtruth Generation.} We first generate groundtruth density maps and then generate groundtruth coarse- and fine-grained attention maps. For density maps, following \cite{zhang2016single}, each labeled head $p_{j}$ is substituted with a Gaussian kernel $\mathcal{N}\left(p_{i}, \sigma^{2}\right)$, where $\sigma$ is the average distance of $p_{j}$ and its 3 nearest neighbors. The kernel is normalized to 1, so the integral of the density map is equal to the labeled people count. For CAM, we set the value to 1 for one pixel if the value $\geq 1e-5$ for corresponding position in density map and 0 otherwise to obtain the groundtruth. For FAM, we categorize one pixel to class 0 if its value $< 1e-5$ in density map, and then count all the other values and divide them into $k-1$ categories from large to small, and each pixel’s class is assigned accordingly.

We adopt Mean Absolute Error(MAE) and Root Mean Square Error(RMSE) for evaluating the counting accuracy:
\begin{equation}
    MAE=\frac{1}{N}\sum_{i=1}^{N}\left|Count^{i}_{est}-Count^{i}_{gt}\right|
    \label{eq:mae}
\end{equation}
\begin{equation}
    RMSE=\sqrt{\frac{1}{N}\sum_{i=1}^N(Count^{i}_{est}-Count^{i}_{gt})^2}
    \label{eq:rmse}
\end{equation}
SSIM and PSNR are used for evaluating the quality of density maps.

\subsection{Implementation Details}
During training, each image is random cropped to $\frac{1}{2}$ size, then horizontal flipped at possibility 0.5 to enlarge the dataset. Pre-trained VGG-16 on ImageNet is used to initialize the parameters of the feature map extractor, and other parameters are randomly initialized by Gaussian distribution with $\delta=0.01$. We train our network with Adam optimizer for 500 epochs and learning rate is initially set to 2e-5 and reduced by half every 100 epochs. We also multiply the density maps by an expansion factor=50 as the directly generated pixel values are quite small.

\subsection{Comparison with State-of-the-art Methods}
We compare our model with state-of-the-art crowd density maps estimation methods since 2016.

\textbf{ShanghaiTech} dataset consists of PartA and PartB. PartA has 300 training images and 182 testing images with relatively high density. PartB has 400 training images and 316 testing images with relatively low density.
As Table \ref{tab:sh} shows, CFANet outperforms the state-of-the-art methods. Specifically, we improve the MAE by 1.6\%  and achieve the second best RMSE on PartA sub-dataset, and improve the MAE by 3.0\% and the RMSE by 1.0\% on PartB sub-dataset. These results suggest that our method can be well applied in both crowded and sparse scenes.
\begin{table}[ht]
    \centering
    \begin{tabular}{|l|c|c|c|c|}
    \hline
         \multirow{2}{*}{Method}&\multicolumn{2}{c|}{PartA}&\multicolumn{2}{c|}{PartB}  \\
         \cline{2-5}
         & MAE&RMSE&MAE&RMSE\\\hline
         CSRNet\cite{li2018csrnet}&68.2&115.0&10.6&16.0\\\hline
         SANet\cite{cao2018scale}&67.0 &104.5 &8.4 &13.6\\\hline
         TEDNet\cite{jiang2019crowd}&64.2 & 109.1 &8.2 &12.8\\\hline
         ADCrowdNet\cite{liu2019adcrowdnet}&63.2 &98.9 & 7.7&12.9 \\\hline
         AT-CSRNet\cite{zhao2019leveraging}&-&-&8.1&13.5\\\hline
         CAN\cite{liu2019context}&62.3 &100.0 &7.8 &12.2 \\\hline
         Bayesian loss\cite{ma2019bayesian}&62.8 &101.8 &7.7 &12.7 \\\hline
         DSSINet\cite{liu2019crowd}&60.6 &96.0 &6.9&\underline{10.3}\\\hline
         RANet\cite{zhang2019relational}&59.4&102.0&7.9&12.9\\\hline
         S-DCNet\cite{xiong2019open}&58.3&95.0&\underline{6.7}&10.7\\\hline
         PGCNet\cite{yan2019perspective}&\underline{57.0}&\textbf{86.0}  &8.8 &13.7 \\\hline
\textbf{Ours} & \textbf{56.1}&\underline{89.6}&\textbf{6.5}&\textbf{10.2}\\\hline
    \end{tabular}
    \caption{Comparison of performance on ShanghaiTech dataset. Top two performance are highlighted in \textbf{bold} and \underline{underline}.}
    \label{tab:sh}
\end{table}

\textbf{UCF\_CC\_50} dataset includes 50 crowd images with extremely high density. This dataset is quite challenging not only because of the quite limited labeled images, but also because the people count in each image varies from 94 to 4543 with an average of 1279.5. We follow the standard 5-fold cross-validation in \cite{idrees2013multi}. As shown in Table \ref{tab:ucf}, we achieve the best MAE and the second best RMSE performance. Although this dataset has very limited labeled images, proposed CFANet can still get a promising result, which shows its capacity when there is a training sample shortage.
\begin{table}[ht]
    \centering
    \begin{tabular}{|l|c|c|}
    \hline
    Method& MAE&RMSE\\\hline
         CSRNet\cite{li2018csrnet}&266.1&397.5\\\hline
         ADCrowdNet\cite{liu2019adcrowdnet}&257.1&363.5\\\hline
         CAN\cite{liu2019context}&212.2&\textbf{243.7} \\\hline
PGCNet\cite{yan2019perspective} & 244.6& 361.2 \\\hline
Bayesian loss\cite{ma2019bayesian}&229.3&308.2 \\\hline
SANet\cite{cao2018scale}&258.4 &334.9\\\hline
TEDNet\cite{jiang2019crowd}&249.4&354.5\\\hline
DSSINet\cite{liu2019crowd}&216.9&302.4\\\hline
RANet\cite{zhang2019relational}&239.8&319.4\\\hline
S-DCNet\cite{xiong2019open}&\underline{204.2}&301.3\\\hline
\textbf{Ours} &\textbf{203.6}&\underline{287.3}\\\hline
    \end{tabular}
    \caption{Comparison of performance on UCF\_CC\_50 dataset. Top two performance are highlighted in \textbf{bold} and \underline{underline}.}
    \label{tab:ucf}
\end{table}

\textbf{UCF-QNRF} is the most up-to-date large scale crowd image dataset. It has 1535 labeled images in total, with 1201 images for training and 334 images for testing. The people count in each image ranges from 49 to 12865, which makes it quite challenging and diverse. Because the resolutions vary in a large range, we resize the images to ensure that the longer side falls in [1024, 2048] in the training process. In the testing process, we use the original images.

The comparison is summarized in Table \ref{tab: qnrf}, our method delivers the best RMSE, surpassing the second best approach by 1.6\%, and the second best MAE, which is quite close to the best one.

\begin{table}[ht]
    \centering
    \begin{tabular}{|l|c|c|}
    \hline
    Method& MAE&RMSE\\\hline
         CAN\cite{liu2019context}&107&183 \\\hline
        TEDNet\cite{jiang2019crowd}&113&188\\\hline
        RANet\cite{zhang2019relational}&111&190\\\hline
        S-DCNet\cite{xiong2019open}&104.4&176.1\\\hline
        SFCN\cite{wang2019learning}&102.0&171.4\\\hline
        DSSINet\cite{liu2019crowd}&99.1&159.2\\\hline
        MBTTBF-SFCB\cite{sindagi2019multi}&97.5&165.2\\\hline
        Bayesian loss\cite{ma2019bayesian}&\textbf{88.7}&\underline{154.8} \\\hline
        \textbf{Ours} &\underline{89.0}&\textbf{152.3}\\\hline
    \end{tabular}
    \caption{Comparison of performance on UCF-QNRF dataset. Top two performance are highlighted in \textbf{bold} and \underline{underline}.}
    \label{tab: qnrf}
\end{table}

\textbf{Mall} dataset \cite{chen2012feature} consists of 2000 frames obtained from a stationary surveillance camera in a mall. The images have a resolution of 320$\times$480. Following the same setting as \cite{chen2012feature}, the first 800 frames are used as training frames and the remaining 1200 frames are used for testing. It can be viewed from Table \ref{tab: mall} that our method get the best performance in both MAE and RMSE, improving the performance by 6.3\% and 7.1\%, which demonstrates that our method also has the superior capacity in low-density scenes.
\begin{table}[ht]
    \centering
    \begin{tabular}{|l|c|c|}
    \hline
    Method& MAE&RMSE\\\hline
    ConvLSTM\cite{xiong2017spatiotemporal}&2.10&7.6\\\hline
         DRSAN\cite{liu2018crowd}&1.73&2.1 \\\hline
DecideNet\cite{liu2018decidenet}&1.52&1.90 \\\hline
AT-CFCN\cite{zhao2019leveraging}&2.28&2.90\\\hline
E3D\cite{zou2019enhanced}&1.64&2.13\\\hline
SAAN\cite{hossain2019crowd}&\underline{1.28}&\underline{1.68}\\\hline
\textbf{Ours} &\textbf{1.20}&\textbf{1.56}\\\hline
    \end{tabular}
    \caption{Comparison of performance on Mall dataset. Top two performance are highlighted in \textbf{bold} and \underline{underline}.}
    \label{tab: mall}
\end{table}

\begin{figure}[ht]
    \centering
    \includegraphics[width=\linewidth]{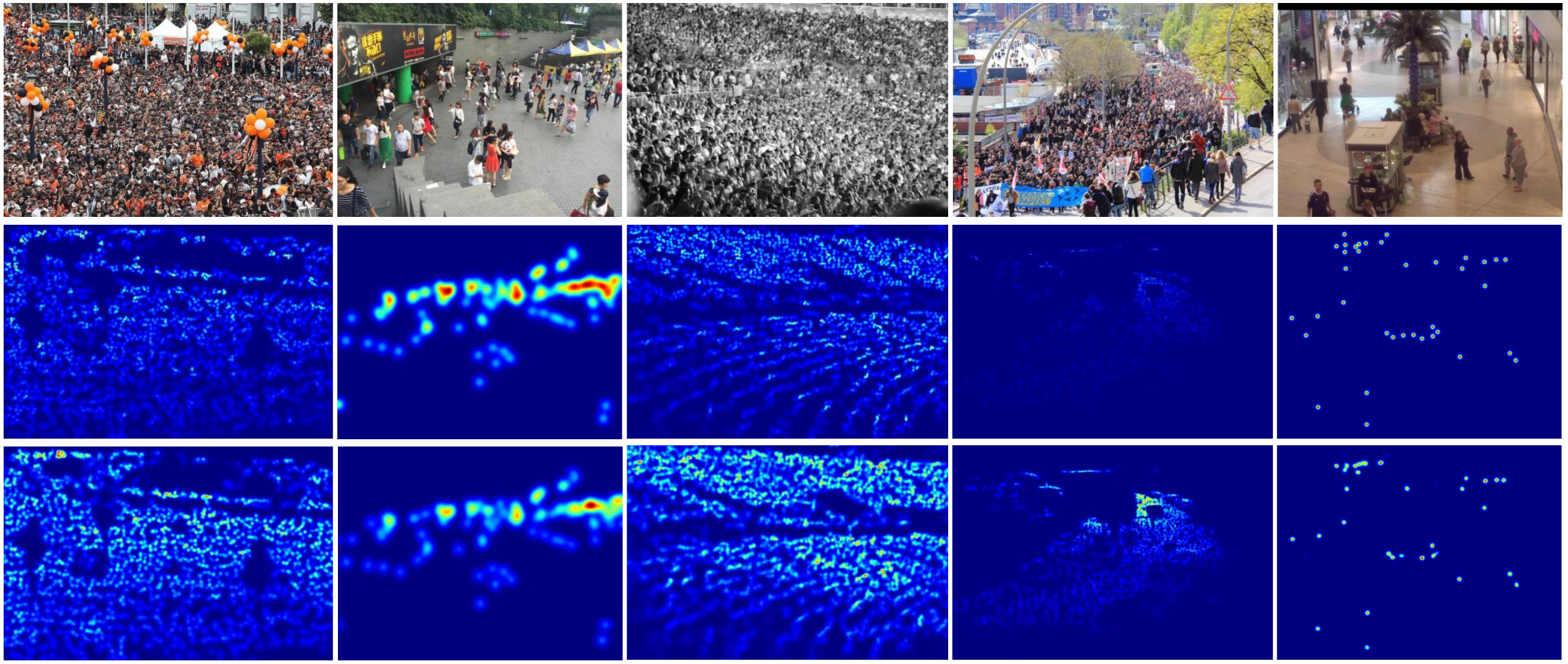}
    \caption{Visualization of density maps. Three rows from top to bottom are input image, groundtruth and estimated density maps from CFANet respectively.}
    \label{fig:results}
\end{figure}

We select representative images from each dataset and compare the predicted density map with groundtruth in Fig. \ref{fig:results}. It can be viewed that the distribution of crowd is very close to the groundtruth.

\begin{figure}[ht]
    \centering
    \includegraphics[width=\linewidth]{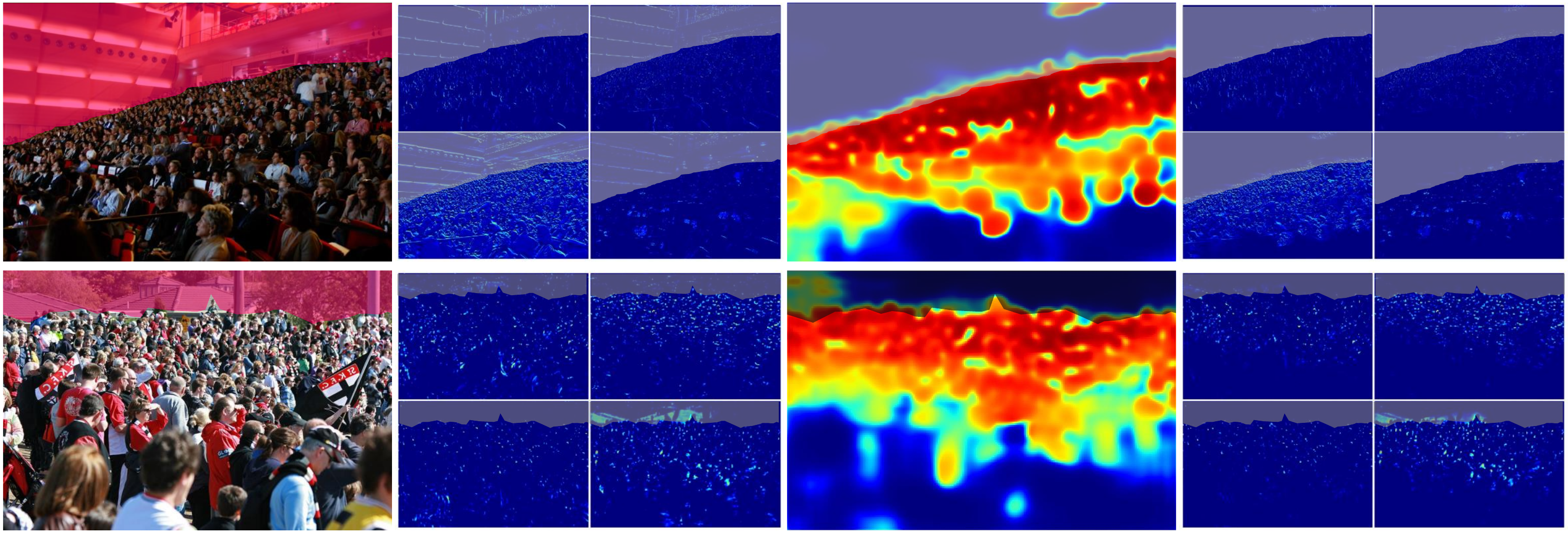}
    \caption{Feature maps before and after refined by fine-grained attention maps. Four columns from left to right are input image, original feature maps, fine-grained attention map (FAM) and feature maps refined by FAM. The background area is highlighted by the mask.}
    \label{fig:fm}
\end{figure}

Some feature maps from internal layers are visualized in \ref{fig:fm}. Refined feature maps in the last column retain most feature in crowd areas and discard unrelated features in background areas and help reduce false recognition ratio.

We also compare the density map quality under SSIM and PSNR criterion. The performances are summarized in Table \ref{tab:psnr}. On the ShanghaiTech PartA dataset, density maps produced by our method have the best quality. Note that our estimated density maps have the same resolution as the input image. This indicates that proposed CFANet has the capacity of improving the quality of high-resolution density maps as well as predicting accurate people count.
\begin{table}[ht]
    \centering
    \begin{tabular}{|l|c|c|}
    \hline
    Method&SSIM&PSNR\\
    \hline
    MCNN\cite{zhang2016single}&0.52&21.40\\\hline
       CSRNet\cite{li2018csrnet}  & 0.76&23.79 \\\hline
        ADCrowdNet\cite{liu2019adcrowdnet} &\textbf{0.88}&24.48\\\hline 
        TEDNet\cite{jiang2019crowd}&0.83&25.88\\\hline
        \textbf{Ours}&\textbf{0.88}&\textbf{30.11}\\
        \hline
    \end{tabular}
    \caption{SSIM and PSNR comparisons on ShanghaiTech PartA dataset.}
    \label{tab:psnr}
\end{table}

As described previously, we propose this method to suppress the influence of irrelevant background and reduce false recognition. To find out if CFANet can function well in this aspect, we set a criterion for judging the ratio of false recognition in the background. In the testing process, we separate the estimated people count in the background and crowd area, and calculate the proportion of these two parts to the total number of people. To be more specific, for each image in the testing set, the pixels with value$< 1e-5$ are considered as in the background area, and others are considered as in the crowd area. By accumulating the pixel values of the background area, we can obtain the corresponding estimated number of people: $Count^{bg}_{est}$. The ratio $r_{bg}=Count^{bg}_{est}/Count_{est}$ represents the false recognition ratio, and the lower it is, the less false recognition there is. We compare the $r_{bg}$ of proposed method with two open-source methods CSRNet\footnote{https://github.com/leeyeehoo/CSRNet-pytorch} and CAN\footnote{https://github.com/weizheliu/Context-Aware-Crowd-Counting} and removing the BL item (Equation \ref{eq:bl}) in loss function. Results in Table \ref{tab:ratio} validate the effectiveness of CFANet and background-aware loss item in suppressing the false recognition in the background area.
\begin{table}[ht]
    \centering
    \begin{tabular}{|l|c|c|}
    \hline
    Method& PartA&PartB\\
    \hline
       CSRNet\cite{li2018csrnet}  & 0.0381 &0.0857\\\hline
        CAN\cite{liu2019context} & 0.0283&0.0359\\\hline
        Ours w/o bg-aware loss&0.0205&0.0245\\\hline
        Ours w. bg-aware loss&\textbf{0.0184}&\textbf{0.0219}\\
        \hline
    \end{tabular}
    \caption{Comparison of false recognition ratio in estimated density maps on ShanghaiTech dataset.}
    \label{tab:ratio}
\end{table}

\subsection{Ablation Study}
We conduct extensive ablation studies on ShanghaiTech PartA dataset to analyze the impact of different settings of the network and training process.

\textbf{The effectiveness of CRR and DLE}. Since we have introduced two new branches: CRR and DLE, we first try to analyze the impact of branch setting and identify if they can bring improvement to the performance. We construct the 'baseline' by removing the CRR and DLE. We then add the CRR and DLE respectively to build 'w. CRR' and 'w. DLE'. Finally, both branches are integrated as 'w. CRR+DLE', which is the default setting in experiments. The results are reported in Table \ref{tab:branch}. We can see that the integration of CRR and DLE branch can both bring improvement, demonstrating the effectiveness of attention mechanism, and by combining them the best performance can be achieved.
\begin{table}[ht]
\centering
\begin{tabular}{|l|c|c|}
\hline
Network & MAE & RMSE \\
\hline
baseline & 63.7&102.9\\\hline
w. CRR & 56.9&91.8\\\hline
w. DLE & 60.2&99.4\\\hline
w. CRR+DLE & \textbf{56.1}&\textbf{89.6}\\
\hline
\end{tabular}
\caption{Ablation study on CRR and DLE branch configuration.}
\label{tab:branch}
\end{table}

\textbf{Multi-level supervision}. Table \ref{tab:lossn} reports the performance of using different supervision's setting. $Supervision\ i$ corresponds using the $i$-th blue arrow in Fig. \ref{fig:cfanet}. We can see that the commonly used setting, single $Supervision\ 4$, can only produce a relatively low accuracy. The more supervision is integrated, the more improvement it can bring to the model's accuracy and using all 4 supervision achieves the best performance. This undoubtedly demonstrates the superiority of the multi-level supervision mechanism.
 
\begin{table}[ht]
\centering
\begin{tabular}{|l|c|c|c|}
\hline
Supervision & MAE & RMSE \\
\hline
$Supervision\ 4$ & 69.5&109.1\\\hline
$Supervision\ 3\sim 4$ & 64.6&106.3\\\hline
$Supervision\ 2\sim 4$ & 62.0&97.1\\\hline
$Supervision\ 1\sim 4$ & \textbf{56.1}&\textbf{89.6}\\
\hline
\end{tabular}
\caption{Ablation study on supervision setting.}
\label{tab:lossn}

\end{table}

\textbf{Class number in DLE branch}. As described previously, we need to specify a class number for the multi-class classification problem in DLE branch. If the class number is too small, the features of each category are not representative enough. If the number is too big, the differences between classes can be too little to capture. To find out an appropriate setting, we have tried setting it to 4, 6, 8, 10, and experimental results in Table \ref{tab:level} show that setting it to 6 can achieve the best performance.
\begin{table}[ht]
\centering
\begin{tabular}{|l|c|c|}
\hline
Class number & MAE & RMSE \\
\hline
4 & 56.9&91.3\\\hline
6 & \textbf{56.1}&\textbf{89.6}\\\hline
8 & 57.4&91.7\\\hline
10 & 57.9&92.9\\
\hline
\end{tabular}
\caption{Ablation study on density level class number setting.}
\label{tab:level}
\end{table}

\section{The Impact of Loss Functions}
Many kinds of loss functions have been proposed to improve the performance of density map estimation models. Here we firstly describe existing loss functions briefly, and then compare proposed BSL with them.

Mean Square Error ($MSE$) is the most widely used loss function which computes the average of the square error of each pixel in density maps. 

Spatial Abstraction Loss ($L_{SA}$) and Spatial Correlation Loss ($L_{SC}$) are proposed by \cite{jiang2019crowd}. $L_{SA}$ computes $MSE$ losses on multiple abstraction levels and $L_{SC}$ further computes global consistency.

Mean Absolute Error ($MAE$) is introduced by \cite{shi2019counting} to add robustness to outliers. $MAE$ and $MSE$ are jointly optimized in training. 

Multi-scale density level consistency loss ($L_c$) is proposed by \cite{dai2019dense} which combines pixel-wise $MSE$ with $MAE$ of globally averaged density maps. 

Bayesian loss ($L^{Bayes+}$) is introduced by \cite{ma2019bayesian}, which constructs a density contribution probability model from the point annotations, instead of constraining the value at every pixel in
the density map.

SSIM is firstly introduced by \cite{cao2018scale} as loss function ($L_{SSIM}$) to improve the quality of results by incorporating the local correlation in density maps. 

Dilated Multiscale Structural Similarity (DMS-SSIM) loss ($L_{DS}$) is introduced by \cite{liu2019crowd}. The DMS-SSIM network for computing the $L_{DS}$ consists of $m=5$ dilated convolutional layers with dilation rates of 1, 2, 3, 6 and 9.

\subsection{Impact on Counting Accuracy}
We evaluate the impact of loss functions on counting accuracy by training CFANet with the above existing loss functions and proposed BSL on ShanghaiTech PartA and UCF-QNRF datasets. As Table \ref{tab:msl_datasets} shows, BSL achieves the best performance on both datasets, which validate its effectiveness.
\begin{table}[ht]
\centering
\begin{tabular}{|l|c|c|c|c|}
\hline
 \multirow{2}{*}{Loss function}&\multicolumn{2}{c|}{ST PartA}&\multicolumn{2}{c|}{UCF-QNRF}\\
\cline{2-5}
 & MAE & RMSE & MAE &RMSE\\
\hline
$MSE$&63.6&104.5&101.9&170.4\\
$L_{SA}+L_{SC}$\cite{jiang2019crowd}&64.2&109.1&113.0&188.0\\
$L_{C}$\cite{dai2019dense}&65.2 &107.7 & 101.3 &174.2\\
$MSE+MAE$\cite{shi2019counting}&59.7 &98.2 &99.4 &164.8\\
$L_{Bayes+}$\cite{ma2019bayesian}&66.4 &113.9 &97.2 &160.6\\
$L_{SSIM}$\cite{cao2018scale}&64.9 &107.4 &108.8 &179.5\\
$L_{DS}$\cite{liu2019crowd}&60.6 &96.0 &99.1 &159.2\\
$BSL$ & \textbf{56.1}&\textbf{89.6}&\textbf{89.0}&\textbf{152.3}\\
\hline
\end{tabular}

\caption{Comparison of loss functions on ShanghaiTech PartA and UCF-QNRF datasets.}
\label{tab:msl_datasets}
\end{table}

In order to further validate BSL's robustness on different models, we replace original $MSE$ loss function with proposed BSL on MCNN\cite{zhang2016single}, CSRNet\cite{li2018csrnet} and CAN\cite{liu2019context} on ShanghaiTech PartA dataset. Results in Table \ref{tab:msl_models} show that, without changing the network, using BSL can improve the performance on multiple models significantly.

\begin{table}[ht]
\centering
\begin{tabular}{|l|c|c|}
\hline
 \multirow{2}{*}{Model}&\multicolumn{2}{c|}{Using MSE/BSL}\\
\cline{2-3}
& MAE & RMSE\\
\hline
MCNN\cite{zhang2016single} & 110.2/\textbf{88.2$\uparrow^{20.0\%}$}&173.2/\textbf{140.5$\uparrow^{18.9\%}$}\\
CSRNet\cite{li2018csrnet} & 68.2/\textbf{63.1$\uparrow^{7.5\%}$}&115.0/\textbf{102.5$\uparrow^{10.9\%}$}\\
CAN\cite{liu2019context} & 62.3/\textbf{59.0$\uparrow^{5.3\%}$}&100.0/\textbf{90.1$\uparrow^{9.9\%}$}\\
\hline
\end{tabular}

\caption{Comparison of BSL and MSE on multiple models.}
\label{tab:msl_models}
\end{table}

\section{Conclusion}
In this paper, we present a new model named Coarse- and Fine-grained Attention Network (CFANet) for high-quality crowd density map generation and people counting. By incorporating Crowd Region Recognizer (CRR) and Density Level Estimator (DLE) to estimate coarse- and fine-grained attention maps, feature maps for regressing the density maps are refined on multi-scale and the network can better focus on the crowd region. We also adopt multi-level supervision to help facilitate the backpropagation of gradient and reduce overfitting. In addition, we propose a novel and effective loss function named Background-aware Structural Loss (BSL), which can achieve better counting accuracy and enhance structural similarity as well as reduce false recognition ratio. Combining proposed CFANet with BSL can outperform current state-of-the-art methods on most mainly used datasets.

{\small
\bibliographystyle{ieee_fullname}
\bibliography{egbib}
}

\end{document}